\documentclass[lettersize,journal]{IEEEtran}

\usepackage{amsmath,amsfonts}
\usepackage{algorithmic}
\usepackage{algorithm}
\usepackage{array}
\usepackage[caption=false,font=normalsize,labelfont=sf,textfont=sf]{subfig}
\usepackage{textcomp}
\usepackage{stfloats}
\usepackage{url}
\usepackage{verbatim}
\usepackage{graphicx}
\usepackage{circledsteps}
\usepackage{fontenc}   
\usepackage{inputenc}   
\usepackage[T1]{fontenc}
\usepackage[italic=false]{mathastext}
\usepackage{xcolor}
\newcommand{\orcidicon}[1]{}
\usepackage[colorlinks=true, citecolor=blue, linkcolor=blue, urlcolor=black]{hyperref}

\makeatletter
\renewcommand\@cite[2]{\textcolor{blue}{[}#1\textcolor{blue}{]}}
\makeatother

\begin{document}

\title{READ-Net: Clarifying Emotional Ambiguity via Adaptive Feature Recalibration for Audio-Visual Depression Detection}

\author{
	Chenglizhao Chen\textsuperscript{1},~~~
	Boze Li\textsuperscript{1},~~~
	Mengke Song\textsuperscript{1}\textsuperscript{\dag},~~~
	Dehao Feng\textsuperscript{1},~~~
	Xinyu Liu\textsuperscript{1},\\
	Shanchen Pang\textsuperscript{1},~~~
	Jufeng Yang\textsuperscript{2},~~~
	Hui Yu\textsuperscript{3}\\[0.3em]
	
	\textsuperscript{1}College of Computer Science and Technology, China University of Petroleum (East China), China\\
	\textsuperscript{2}College of Computer Science, Nankai University, China\\
	\textsuperscript{3}School of Computing Science, University of Glasgow, Scotland, UK

\vspace{-0.7cm}
\thanks{The first two authors contributed equally to this work.}
\thanks{\textsuperscript{\dag}Corresponding author: Mengke Song (songsook@163.com).}
}

\markboth{XXX XXX XXX XXX, VOL.XX, NO.XX, XXX.XXXX}%
{Shell \MakeLowercase{\textit{et al.}}: A Sample Article Using IEEEtran.cls for IEEE Journals}

\maketitle

\begin{abstract}
   Depression is a severe global mental health issue that impairs daily functioning and overall quality of life. Although recent audio-visual approaches have improved automatic depression detection, methods that ignore emotional cues often fail to capture subtle depressive signals hidden within emotional expressions. Conversely, those incorporating emotions frequently confuse transient emotional expressions with stable depressive symptoms in feature representations, a phenomenon termed \emph{Emotional Ambiguity}, thereby leading to detection errors. To address this critical issue, we propose READ-Net, the first audio-visual depression detection framework explicitly designed to resolve \emph{Emotional Ambiguity} through Adaptive Feature Recalibration (AFR). The core insight of AFR is to dynamically adjust the weights of emotional features to enhance depression-related signals. Rather than merely overlooking or naively combining emotional information, READ-Net innovatively identifies and preserves depressive-relevant cues within emotional features, while adaptively filtering out irrelevant emotional noise. This recalibration strategy significantly clarifies feature representations, and effectively mitigates the persistent challenge of emotional interference. Additionally, READ-Net can be easily integrated into existing frameworks for improved performance.
   Extensive evaluations on three publicly available datasets show that READ-Net outperforms state-of-the-art methods, with average gains of 4.55\% in accuracy and 1.26\% in F1-score, demonstrating its robustness to emotional disturbances and improving audio-visual depression detection.
\end{abstract}


\section{Introduction}

\begin{figure}[!t]
    \centering
    \includegraphics[width=1.0\linewidth]{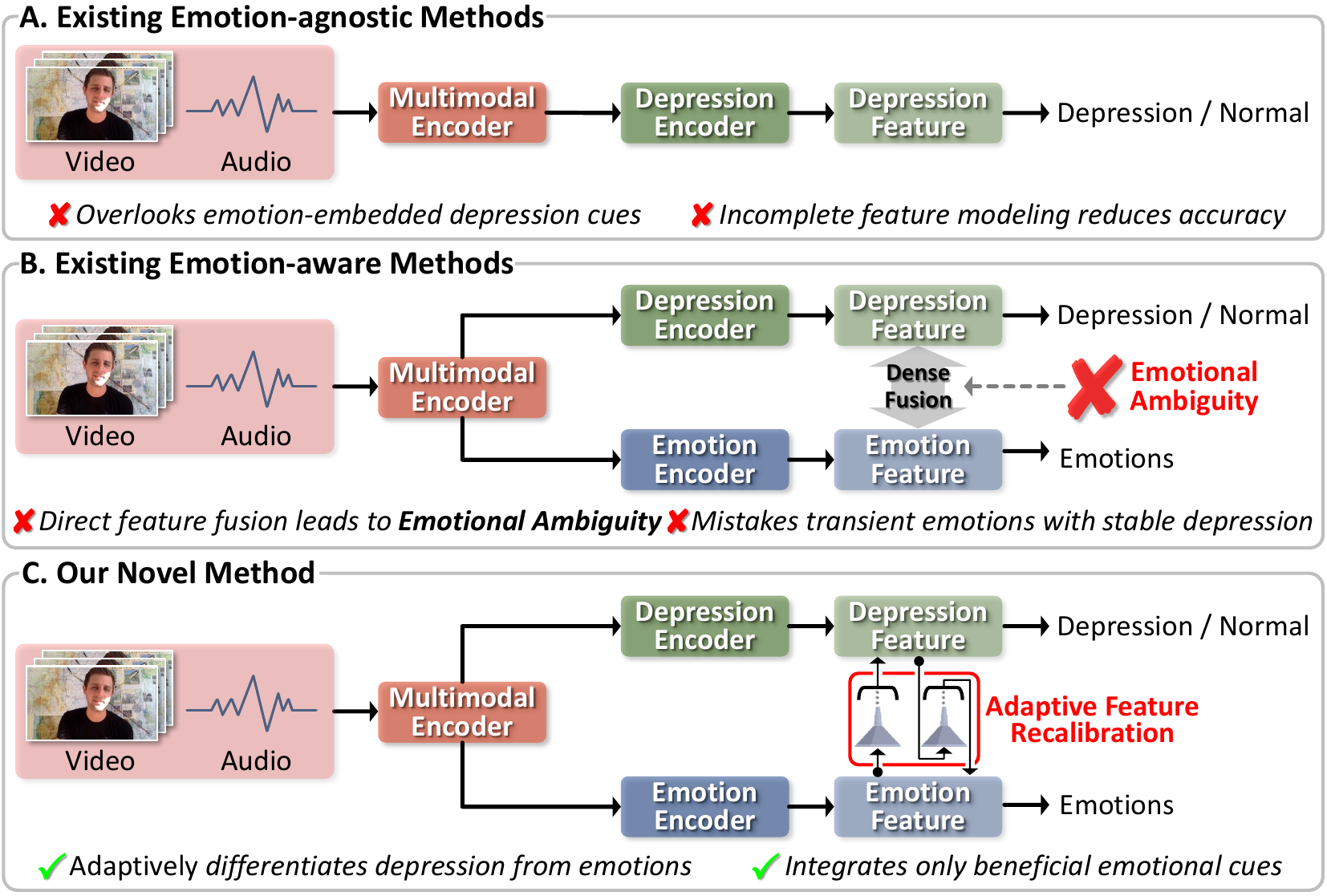}  
    \vspace{-0.6cm}
    \caption{Comparison of existing and proposed methods for audio-visual depression detection. (A) Emotion-agnostic methods overlook emotion-embedded depressive cues, leading to incomplete modeling. (B) Directly fused emotional and depressive features risk confusing transient moods with stable depressive signals. (C) Our proposed approach employs Adaptive Feature Recalibration to selectively integrate beneficial emotional cues, enhancing both accuracy and robustness.}
    \label{fig:motivation}
    \vspace{-0.6cm}
\end{figure}

\IEEEPARstart{D}{epression}, one of the most severe public health challenges in the 21st century, affects more than 280 million people globally. The World Health Organization (WHO) projects that it will become the leading contributor to global disease burden by 2030. Traditional clinical diagnosis mainly depends on subjective tools such as self-report scales and structured interviews, which inherently suffer from limited timeliness, high subjectivity, and susceptibility to patients' concealment of symptoms~\cite{3b2}. 

To address these drawbacks, recent studies have increasingly emphasized \emph{automated depression detection} based on objective biomarkers. Audio-visual modalities, including facial expressions, vocal intonation~\cite{7b1}, and body gestures, have received extensive attention due to their capacity to capture behavioral and physiological cues associated with depressive states~\cite{23a}. Advances in deep learning have driven a shift from unimodal analysis toward multimodal fusion frameworks that integrate complementary sources to model complex affective patterns. However, many multimodal approaches still struggle with simplistic fusion strategies and difficulty in separating transient emotional fluctuations from persistent depressive indicators, which restricts the diagnostic validity and generalizability of current methods. Existing findings further show that depressed individuals display characteristic patterns in facial behavior, vocal attributes, and body movements.


\begin{figure}[!t]
	\centering
	\includegraphics[width=1.0\linewidth]{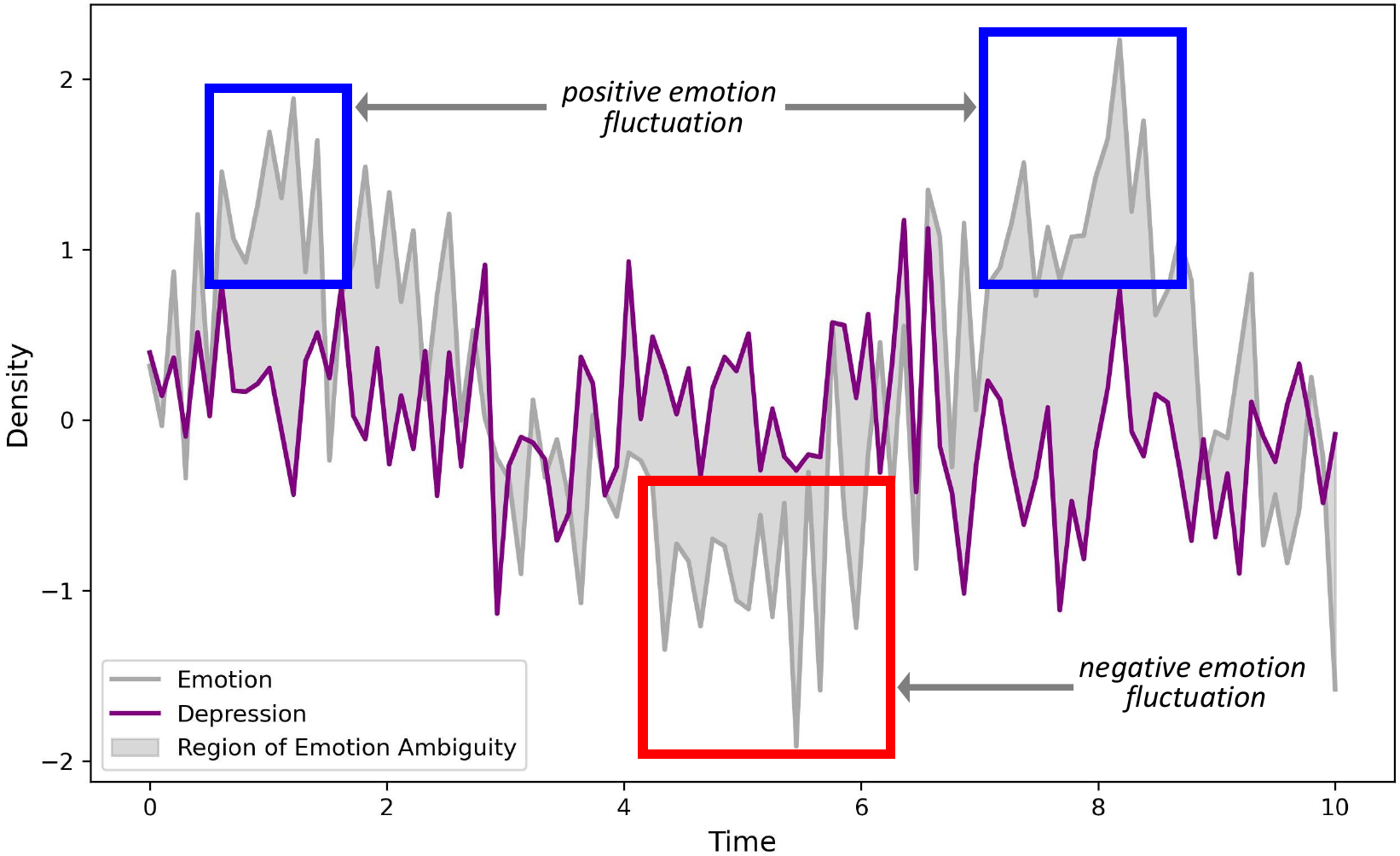}
	\vspace{-0.6cm}
	\caption{Visualization of Emotional Ambiguity. This figure illustrates the overlap between transient emotional fluctuations (positive and negative) and stable depressive signals over time. The shaded region represents the area of \emph{Emotional Ambiguity}, where emotional fluctuations complicate the accurate identification of depressive symptoms.}
	\label{fig:EmotionalAmbiguity}
	\vspace{-0.6cm}
\end{figure}

Current audio-visual depression detection frameworks are generally categorized into emotion-agnostic and emotion-aware methods. Emotion-agnostic approaches~\cite{depmstat,liu2024mddmamba,d1,d2} (Figure~\ref{fig:motivation}-\textbf{A}) exclude emotional cues and focus on depression-specific biomarkers such as speech pauses, gaze aversion, and flattened affect. While avoiding emotional noise, they also discard important information, since depression often alters emotional reactivity, including blunted responses to positive stimuli or context-inappropriate expressions such as brief smiles during distress. Ignoring these emotion-modulated behaviors leads to incomplete feature representations.

Emotion-aware methods~\cite{zheng2023two,teng2024multi,emo2} (Figure~\ref{fig:motivation}-\textbf{B}) explicitly incorporate emotional features under the assumption that depression modifies typical affective patterns. Although such methods capture depression-related emotional fluctuations, they introduce \emph{Emotional Ambiguity} (Figure~\ref{fig:EmotionalAmbiguity}), where short-term emotions, such as momentary sadness or happiness during interviews, are confused with persistent depressive symptoms like chronic emotional numbness. Naive fusion lets transient affect contaminate stable depressive cues, leading to errors, such as misclassifying a healthy individual's brief sadness as depression or masking a depressed individual's state due to occasional positive affect, which ultimately degrades representation quality and generalization.

A key unresolved challenge is distinguishing emotional expressions that reflect depressive pathology from those driven by contextual factors. Real-world emotional responses are shaped by interview content, social interactions, and environmental stimuli, often producing transient affective states unrelated to clinical depression. For example, discussing a joyful memory may elicit happiness, whereas recalling a difficult event may induce sadness unrelated to pathology. This context dependence raises the question of whether models are detecting depressive traits or momentary reactions. Addressing this requires understanding emotion and depression as dynamically intertwined rather than simply additive signals. Otherwise, context-driven emotions may be misinterpreted as depression, reducing interpretability and diagnostic validity.

To mitigate \emph{Emotional Ambiguity}, we propose READ-Net, a novel audio-visual framework that adaptively recalibrates emotion-depression interactions (Figure~\ref{fig:motivation}-\textbf{C}). READ-Net neither removes emotional information nor integrates it indiscriminately. Instead, it dynamically differentiates depression-relevant emotional cues from transient emotional noise via Adaptive Feature Recalibration. This recalibration corrects the limitations of emotion-agnostic designs and prevents conflation of fleeting emotions with stable depressive traits, thereby improving accuracy and generalizability.

Experiments on three benchmark datasets, LMVD~\cite{lmvd}, D-vlog~\cite{dvlog}, and DAIC-WOZ~\cite{avec2017}, demonstrate that READ-Net surpasses state-of-the-art methods, while exhibiting strong robustness to Emotional Ambiguity. Its modular architecture also enables seamless integration into existing frameworks, providing a broadly applicable enhancement for depression detection systems.

In summary, the main contributions of this paper include:
\begin{enumerate}
	\item As the first work in this direction, we formalize the concept of \emph{Emotional Ambiguity} in depression detection and empirically show its negative influence on existing audio-visual methods.
	\item We propose READ-Net, a unified audio-visual depression detection framework that addresses \emph{Emotional Ambiguity} through an Adaptive Feature Recalibration mechanism. The framework can be integrated into existing systems as a plug-and-play module.
	\item We further introduce Hierarchical Feature Separation and Dual Consistency Regularization, which disentangle depression-relevant features from emotional and noise representations.
	\item Experiments on public datasets show that READ-Net outperforms state-of-the-art methods with average improvements of 4.55\% in accuracy and 1.26\% in F1-score, demonstrating strong resilience to Emotional Ambiguity. Codes, datasets, and results are available\footnote{\url{https://github.com/READNet-DemoCode/READ-Net}}. 
\end{enumerate}

\section{Related Works}
\label{sec:related_work}

Automatic depression detection (ADD) has progressed rapidly with audio-visual and increasingly linguistic signals. Benchmarks such as AVEC'13/14 and AVEC'17 established protocols for affect and depression estimation and catalyzed multimodal pipelines~\cite{10.1145/2512530.2512533,10.1145/2661806.2661807,avec2017}. Beyond controlled interviews, in-the-wild resources such as D-vlog~\cite{dvlog} and large-scale datasets like LMVD~\cite{lmvd} improve robustness under real-world conditions, while semi-structured clinical corpora support screening-oriented research~\cite{av2}. Standard acoustic toolkits (openSMILE~\cite{10.1145/2729095.2729097}) and early deep audio models (DepAudioNet~\cite{10.1145/2988257.2988267}) provide strong baselines, and given the prevalence of clinical scales, label reliability concerns such as BDI motivate careful protocol design~\cite{3b2}.

\subsection{Emotion-agnostic Audio-visual Methods}

Many studies model paralinguistic and nonverbal markers of depression without explicit emotion supervision. On the acoustic side, CNN and LSTM encoders on spectrograms or raw waveforms, along with spatiotemporal stacks, capture prosodic and articulation cues of depressive speech~\cite{7b1,10.1145/2988257.2988267}. Self-supervised speech encoders such as wav2vec~2.0 and HuBERT further improve robustness under limited labels~\cite{9585401}. On the visual side, deep networks model facial appearance, optical flow, 3D dynamics, and multi-scale temporal differences to detect diminished expressivity and altered micro-dynamics~\cite{19b2,d1,10262153}. Body-level signals, including self-adaptors and fidgeting, provide complementary cues~\cite{9506822}. Multimodal variants fuse audio-visual streams using hierarchical or bridge-style strategies while remaining emotion-agnostic~\cite{18a2,18a1}. These pipelines are effective but may miss higher-level affective context and nuanced depressive patterns.

\subsection{Emotion-aware Audio-visual Methods}

To capture depression-affect interactions, emotion-aware methods incorporate emotion supervision using auxiliary labels, intermediate affect representations, or attention to emotionally salient segments. Temporal transformers jointly learn depression and emotion~\cite{zheng2023two}, and multi-task frameworks integrate sentiment or affect cues in multimodal fusion~\cite{teng2024multi}. Some approaches explicitly track emotional changes across modalities to stabilize predictions~\cite{40a}. However, uniformly leveraging emotional cues may conflate transient affect such as brief joy or sadness with stable depressive traits like anhedonia, producing \emph{Emotional Ambiguity}. Evidence arises from uncertainty modeling for affect annotation~\cite{19a} and modality-specific observations in depressed facial behavior~\cite{depface}. In speech-text settings, cross-modal attention can bias depression features if contextual or stylistic information is not disentangled~\cite{att2}. Consequently, naive emotion integration may add noise and weaken robustness, especially for long weakly labeled sequences in interviews and vlogs.

\subsection{Feature Separation and Ambiguity Mitigation}

Recent research moves toward \emph{selective} mechanisms that emphasize depression-related evidence while suppressing distractors. In visual modeling, dual attention and element-wise recalibration highlight informative channels and temporal regions. In speech and cross-modal domains, hierarchical and monotonic attention and additive cross-modal alignment localize salient acoustic segments and align audio-text~\cite{att1,att2}. For long-form vlogs, time-aware and joint-attention fusion improve aggregation under weak supervision~\cite{tamfn,jamfn,depmstat}. To scale to long contexts, state-space and long-range temporal models such as Mamba variants support efficient cross-modal fusion~\cite{ye2025depmamba,liu2024mddmamba}. Causal and fair fusion approaches also tackle confounding and demographic bias~\cite{fairrefuse}.

Building on these directions, we explicitly target \emph{Emotional Ambiguity} using ambiguity-aware gating, adaptive feature recalibration, and cross-modal alignment~\cite{liu2024privacy}. The design preserves depression-relevant emotional evidence while suppressing context-driven transient affect and scales to long, weakly labeled sequences common in AVEC-style and vlog data~\cite{10.1145/2512530.2512533,avec2017,dvlog,lmvd}. It integrates the robustness of emotion-agnostic cues~\cite{7b1,18a1,18a2}, the selectivity of attention and recalibration mechanisms~\cite{19b1,att1,att2,liu2022interaction}, and the temporal sensitivity of time-aware fusion~\cite{tamfn,depmstat}, achieving finer-grained and more disentangled multimodal ADD.

\begin{figure*}[!t]
	\centering
	\includegraphics[width=1.0\linewidth]{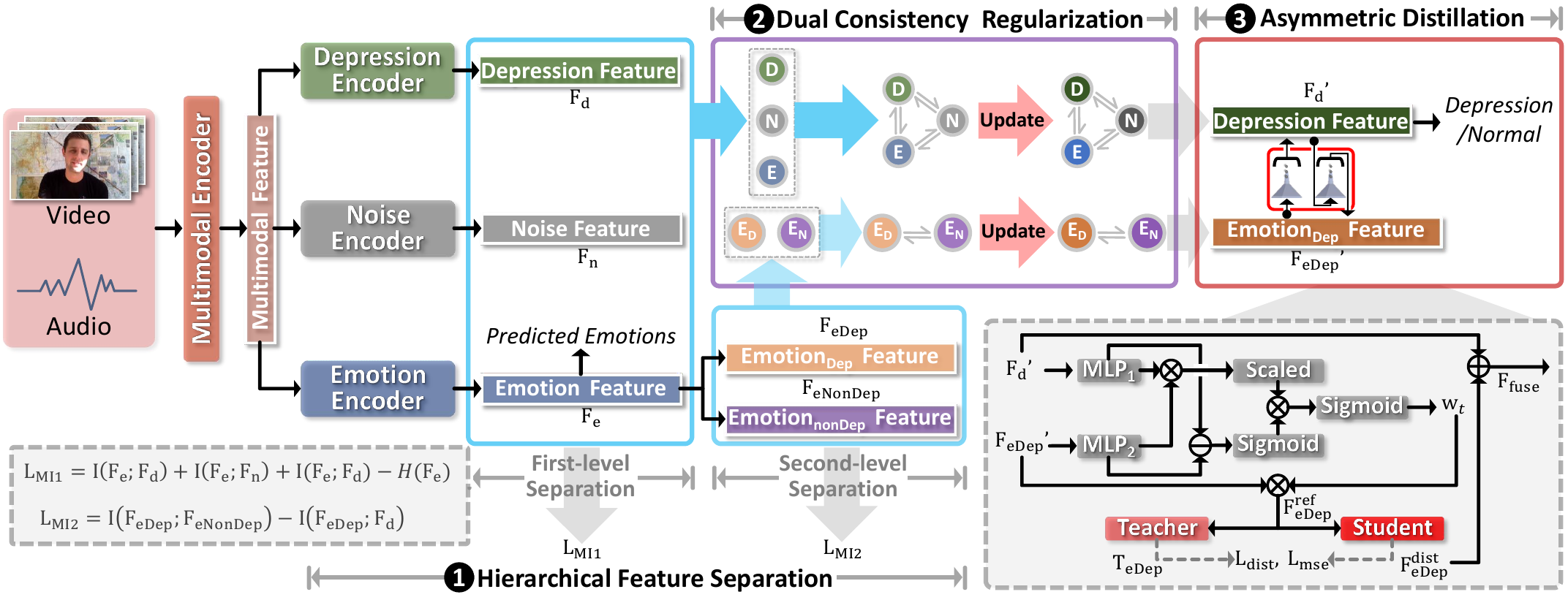}
	\vspace{-0.55cm}
	\caption{Overview of READ-Net. It integrates hierarchical feature separation (\Circled{1}), dual consistency regularization (\Circled{2}), and asymmetric distillation (\Circled{3}) to effectively manage Emotional Ambiguity. By dynamically recalibrating emotion-related features and leveraging multimodal data, READ-Net significantly enhances the accuracy and reliability of depression detection, outperforming traditional methods.}
	\label{fig:pipeline}
	\vspace{-0.3cm}
\end{figure*}

\section{The Proposed Method}
\label{sec:method}

\subsection{Delving Deeper Into Emotional Ambiguity}

Emotional Ambiguity denotes the difficulty of separating transient emotional states from stable depressive symptoms in audio-visual depression analysis. This arises because behavioral cues such as facial expressions, vocal intonation, and body gestures often overlap between momentary emotions and clinical depression. While emotions are dynamic and context-dependent, depression reflects enduring and persistent patterns. Directly integrating emotional features into depression models can mix short-term fluctuations with long-term depressive signals, adding noise and reducing accuracy and generalizability. To clarify this issue, we highlight two representative scenarios:

\textbf{Transient Sadness in Non-Depressed Individuals:}
As shown in Fig.~\ref{fig:EmotionalAmbiguity} (red box), a non-depressed individual may briefly display sadness triggered by external factors, which could appear as a downturned mouth, slower speech, or reduced energy. These cues resemble depressive indicators and may lead to false positives if misinterpreted.

\textbf{Momentary Positive Emotion in Depressed Individuals:}
Conversely, as shown in Fig.~\ref{fig:EmotionalAmbiguity} (blue box), a depressed individual may exhibit brief happiness or engagement, such as a smile during interaction. Such fleeting positive expressions can conceal the underlying depressive state and may result in false negatives when overemphasized.

In both scenarios, momentary emotional expressions can conflict with the subject's true clinical condition. These mismatches are where Emotional Ambiguity emerges for depression detection models.

\textbf{Formalization of Emotional Ambiguity.}
Let $Y \in \{0,1\}$ denote the sequence-level depression label, with
$Y=1$ for depressed and $Y=0$ for non-depressed subjects. Let
$\mathbf{E}_t \in \mathbb{R}^{d_e}$ be the emotion feature at time step $t$, and
$\mathbf{E}_{1:T}=\{\mathbf{E}_1,\dots,\mathbf{E}_T\}$ the emotional trajectory. We consider an emotion-only classifier that outputs the posterior probability $p_\theta(Y \mid \mathbf{E}_t)$ using only emotional information.

Emotional Ambiguity arises when a momentary emotional expression strongly favors the \emph{opposite} depression label relative to the true state. Given a margin $\tau \in (0,0.5)$, the set of emotionally ambiguous time steps is defined as
\begin{equation}
	\mathcal{A}_\tau
	=
	\Big\{
	t \in \{1,\dots,T\} :
	p_\theta(Y = 1-Y \mid \mathbf{E}_t)\ge 0.5+\tau
	\Big\},
	\label{eq:ambiguous-set}
\end{equation}
where $1-Y$ denotes the opposite class. Thus, $\mathcal{A}_\tau$ contains moments when an emotion-only classifier is strongly inclined toward an incorrect label. Based on $\mathcal{A}_\tau$, the sequence-level Emotional Ambiguity rate is
\begin{equation}
	\mathrm{EA}_{\text{err}}(\mathbf{E}_{1:T}, Y)
	=
	\frac{1}{T}
	\sum_{t=1}^{T}
	\mathbb{I}[t \in \mathcal{A}_\tau],
	\label{eq:ambiguity-error}
\end{equation}
where $\mathbb{I}[\cdot]$ is the indicator function. Higher $\mathrm{EA}_{\text{err}}$ indicates that transient emotional expressions frequently mislead the classifier. As illustrated in Fig.~\ref{fig:EmotionalAmbiguity}, this includes: (i) non-depressed subjects exhibiting brief negative or blunted affect that resembles depressive behavior, and (ii) depressed subjects showing temporary positive or neutral affect that masks their condition. Such segments corrupt feature representations and cause unreliable predictions.

The challenge arises because emotional expressions shift rapidly with context, whereas depression is persistent and pervasive. When emotional features are used indiscriminately, transient affective fluctuations contaminate stable depression-related biomarkers (e.g., chronic blunting or gaze aversion), degrading representations and hindering reliable audio-visual depression detection by obscuring the distinction between persistent depressive signals and temporary emotional noise.

\begin{algorithm}[t] 
	\caption{Training and Inference Pipeline of READ-Net}
	\label{alg:readnet}
	\begin{algorithmic}[0]  
		\REQUIRE Dataset $\mathcal{D}=\{(x_i^a,x_i^v,y_i)\}$
		\ENSURE Prediction $\hat{y}$
		
		\vspace{4pt}
		\STATE \textbf{Stage 0: Pre-training and Extraction}
		\STATE Train an emotion recognition branch to obtain extractor $\mathcal{E}_{emo}$.
		\STATE Extract raw audio--visual features $F^{raw}$ for each sample.
		
		\vspace{4pt}
		\STATE \textbf{Stage 1: Hierarchical Feature Separation (HFS)}
		\STATE Encode $F^{raw}$ via three streams to obtain $F_d$, $F_n$, and $F_e$.
		\STATE Compute first-level MI loss $L_{\text{MI1}}$.
		\STATE Decompose $F_e\!\rightarrow\!\{F_{e}^{Dep},F_{e}^{NonDep}\}$ and compute $L_{\text{MI2}}$.
		
		\vspace{4pt}
		\STATE \textbf{Stage 2: Dual Consistency Regularization (DCR)}
		\STATE Build temporal graphs on $\{F_d, F_n, F_e\}$ and update features via GCN.
		\STATE Compute graph consistency loss $L_{\text{graph}}$.
		\STATE Build child graphs on $\{F_{e}^{Dep},F_{e}^{NonDep}\}$ and compute $L_{\text{flow}}$.
		
		\vspace{4pt}
		\STATE \textbf{Stage 3: Asymmetric Distillation (AD)}
		\STATE Apply attention to refine $F_{e}^{Dep}$ to $F_{e}^{ref}$.
		\STATE Distill $F_{e}^{ref}$ to $F_{e}^{dist}$ via teacher--student GRUs.
		\STATE Fuse $F_d$ and $F_{e}^{dist}$ to obtain $F_{\text{fuse}}$.
		
		\vspace{4pt}
		\STATE \textbf{Stage 4: Optimization}
		\STATE Predict depression score $\hat{y}$ from $F_{\text{fuse}}$.
		\STATE Compute $L_{\text{cls}}$ and total loss $L_{\text{total}}$.
		\STATE Update all parameters by back-propagation.
		
		\vspace{4pt}
		\STATE \textbf{Inference}
		\STATE Use the pre-trained $\mathcal{E}_{emo}$ and trained READ-Net to run Stages~1--3 in forward mode to obtain $\hat{y}$.
	\end{algorithmic}
\end{algorithm}

\subsection{Our Method to Clarify Emotional Ambiguity}
To address Emotional Ambiguity, READ-Net adopts an Adaptive Feature Recalibration mechanism~\cite{19b1,19b2} that evaluates the relevance of emotional features to depression, preserving depression-related cues such as blunted affect while suppressing situational emotional noise.

Adaptive Feature Recalibration (Sec.~\ref{sec:hfs}) contains three components. Hierarchical Feature Separation integrates emotional features with depression features by retaining depression-related cues and avoiding direct fusion of transient emotional fluctuations. Dual Consistency Regularization (Sec.~\ref{sec:dcr}) optimizes the relationship between emotional and depression features, ensuring that only persistent emotion patterns aligned with depressive traits are kept. Asymmetric Distillation (Sec.~\ref{sec:ad}) further strengthens depression-relevant emotional cues by selectively distilling useful information from mixed emotional signals, reducing noise influence and improving prediction precision.

Together, these components enable READ-Net to distinguish depression-related emotional cues from transient emotions and improve robustness. Algorithm~\ref{alg:readnet} summarizes the pipeline: Stage~0 pre-trains the emotion branch and extracts audio-visual features; Stages~1--3 apply HFS, DCR, and AD to disentangle and fuse depression-relevant representations, producing $F_{\text{fuse}}$ and prediction $\hat{y}$; Stage~4 jointly optimizes all modules using the total loss $L_{\text{total}}$, and inference uses the trained modules without updates.

As shown in Algorithm~\ref{alg:readnet}, each stage directly corresponds to the modules detailed in Secs.~\ref{sec:hfs}--\ref{sec:ad}. Stage~1 implements the Hierarchical Feature Separation module with the mutual-information-based objectives $L_{\text{MI1}}$ and $L_{\text{MI2}}$. Stage~2 performs Dual Consistency Regularization, introducing the graph-based losses $L_{\text{graph}}$ and $L_{\text{flow}}$ to enhance intra-type coherence and suppress cross-type interference. Stage~3 realizes Asymmetric Distillation, which refines and distills depression-related emotion features into the depression branch. Stage~4 then combines the classification loss $L_{\text{cls}}$ with all regularization terms into the total training loss $L_{\text{total}}$ used in the subsequent experiments.

\subsection{Hierarchical Feature Separation}
\label{sec:hfs}

As shown in Fig.~\ref{fig:pipeline}~(\Circled{1}), multimodal depression detection data such as audio and video often contain both depression-related emotional cues and irrelevant emotional fluctuations. The key challenge is separating depression-specific signals from noise or non-depression emotions. Without effective disentanglement, models may misinterpret transient emotions or fail to capture true depressive states. To address this, we introduce Hierarchical Feature Separation, which isolates depression-relevant features from emotional noise and improves prediction precision.

The central idea is to first extract the overall emotional signal from multimodal data and then further distinguish depression-related emotional cues from general emotional variations. This two-level separation is essential. The first level isolates depression-specific patterns from general emotional and noisy components, enabling the model to concentrate on relevant signals. The second level refines this by separating depression-related emotions from irrelevant ones so that subtle non-depressive emotions do not interfere with detection. This hierarchical strategy enhances the model’s focus on core depressive evidence while filtering transient signals.

\subsubsection{Feature Extraction}
Given multimodal input \( F_{raw} \in \mathbb{R}^{T \times D} \), we construct \emph{three parallel streams} as independent feature extractors operating on the same evidence but serving distinct roles. The \textbf{depression stream} produces \(F_d\) through a multimodal architecture combining multiscale temporal convolution networks with bidirectional cross-modal attention, followed by hierarchical temporal aggregation to obtain rich depression-related temporal-spectral representations.

The \textbf{noise stream} estimates \(F_n\) using the same backbone but with a different training objective. Gradient reversal layers and an orthogonality constraint encourage the representation to diverge from the depression manifold. Specifically, we impose
\( L_{orth} = |\langle F_d, F_n \rangle|^2 \),
which minimizes the inner product between \(F_d\) and \(F_n\) so that \(F_n\) spans an orthogonal feature space.

The \textbf{emotion stream} generates \(F_e\) using the same network structure as the depression branch but with a six-class emotion head (happiness, sadness, anger, fear, surprise, disgust), capturing general emotional expressions for subsequent assessment.

\subsubsection{Pretraining and Knowledge Transfer}
To integrate emotion features effectively into the depression detection model, we began by pretraining the emotion recognition branch using carefully annotated datasets. On the DAIC-WOZ dataset, we labeled six basic emotions for each frame using facial action units (AUs)~\cite{908962}. After feature extraction, filtering, and standardization, we created a balanced emotion training set with 1,200 samples. To address the variability across datasets (D-vlog lacking facial AUs, and LMVD missing original audio data), we extracted compatible emotion features from DAIC-WOZ, ensuring consistency across datasets.

We employed a two-stage training strategy: first, we pretrained the emotion recognition branch independently, achieving over 70\% validation accuracy. Next, we froze its parameters and used it as a fixed feature extractor for the depression detection task. This emotion knowledge transfer mechanism enhanced depression detection across structurally diverse datasets by capturing emotion expression patterns specific to depression. 

\subsubsection{First-Level Separation}
The first-level separation aims to disentangle depression features, noise features, and emotion features by maximizing the mutual information between these feature sets. This step ensures that the model can isolate the relevant depression-related features and remove noise or irrelevant emotional signals.
The loss function is defined as:
\begin{equation}
	L_{MI1} = \lambda_0 \cdot I(F_d; F_n) + \lambda_1 \cdot I(F_e; F_d) + \lambda_2 \cdot I(F_e; F_n) - \lambda_3 \cdot H(F_e)
\end{equation}
where \(I(\cdot;\cdot)\) represents the neural mutual information estimator, \(H(\cdot)\) denotes the entropy regularization term, and \(\lambda_0, \lambda_1, \lambda_2, \lambda_3\) are weighting hyperparameters.

\subsubsection{Second-Level Separation}
Once the emotion features \( F_e \) have been separated at the first level, the second-level separation focuses on further isolating depression-related emotional features \( F_{eDep} \) from non-depression-related emotional features \( F_{eNonDep} \). This step enhances the model's specificity by ensuring transient emotions don't interfere with recognizing stable depressive features.

The mutual information is computed between depression-related emotion features \( F_{eDep} \) and depression features \( F_d \), while minimizing the mutual information between depression-related emotion features and non-depression-related emotion features \( F_{eNonDep} \). The loss function is defined as:
\begin{equation}
	L_{MI2} = I(F_{eDep}; F_{eNonDep}) - I(F_{eDep}; F_d)
\end{equation}

By combining hierarchical feature separation, emotion pretraining, and knowledge transfer, we ensure that depression detection models can effectively isolate depression-relevant features while handling Emotional Ambiguity, leading to more accurate and robust performance across a variety of datasets.

\begin{figure}[!t]
	\centering
	\includegraphics[width=1.0\linewidth]{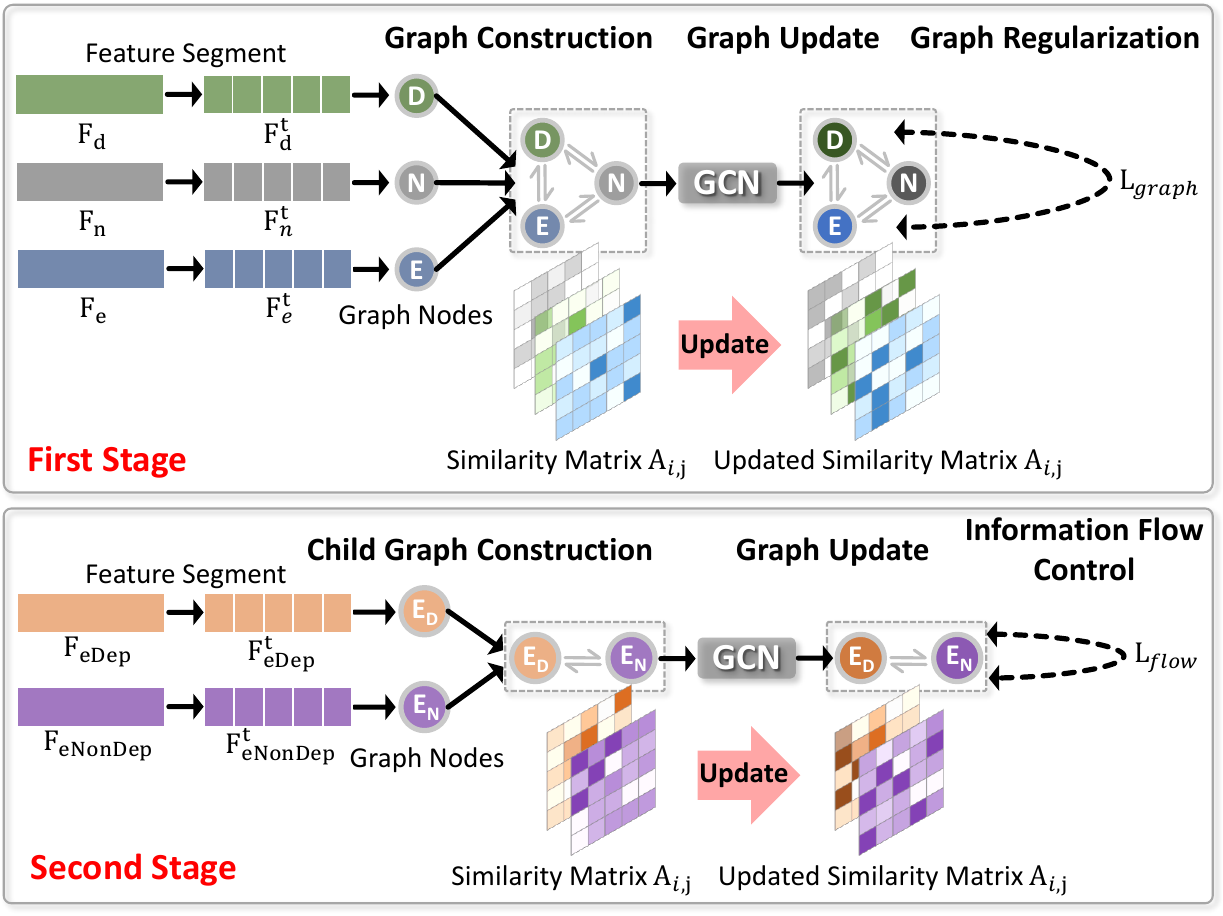}
	\vspace{-0.6cm}
	\caption{Dual Consistency Regularization (DCR) Workflow. Illustrates DCR’s two stages: the first stage separates \( F_d \), \( F_n \), and \( F_e \) via graph construction, updates, and regularization (\( L_{\text{graph}} \)); the second stage refines \( F_{eDep} \) and \( F_{eNonDep} \) using child graph construction, updates, and information flow control (\( L_{\text{flow}} \)), enhancing discriminability and robustness.}
	\label{fig:DualConsistencyRegularization}
	\vspace{-0.6cm}
\end{figure}

\subsection{Dual Consistency Regularization}
\label{sec:dcr}
As illustrated in Figure~\ref{fig:pipeline}~(\Circled{2}), Dual Consistency Regularization (DCR) operates on the outputs of the Hierarchical Feature Separation (HFS) module. HFS initially disentangles raw multimodal features into depression features \(F_d\), noise features \(F_n\), and emotion features \(F_e\), followed by a refinement of \(F_e\) into depression-related \(F_{eDep}\) and non-depression-related \(F_{eNonDep}\). However, these features often retain subtle interdependencies due to overlapping temporal dynamics and complex multimodal interactions, which can undermine separation clarity and model robustness. 

To address this, we introduce DCR, a batch-optimized method to enhance feature discriminability by enforcing consistency within feature types and minimizing cross-type correlations. As shown in Figure~\ref{fig:DualConsistencyRegularization}, DCR operates in two stages: dynamic graph construction and regularization (Sec.~\ref{sec:DynamicGraphConstruction}), and child feature dynamic regularization (Sec.~\ref{sec:Child}), ensuring precise separation while maintaining computational efficiency.

\subsubsection{Dynamic Graph Construction and Regularization}
\label{sec:DynamicGraphConstruction}
As shown in Figure~\ref{fig:DualConsistencyRegularization}-First Stage, this stage leverages temporal relationships across features \( F_d \), \( F_n \), and \( F_e \), each represented as batch tensors \( \mathbb{R}^{T \times D} \), where \( T \) is sequence length, and \( D \) is the feature dimension. 

\textbf{Graph Construction.}
To capture temporal dynamics, these features are treated as nodes in dynamic graphs, with each sequence split into \( K \) segments (e.g., \( K = 5 \)). Each segment corresponds to a graph node \( V_t = \{ F_d^t, F_n^t, F_e^t \}, t = 1, \ldots, K \), enabling us to efficiently capture dynamics at different time steps for the batch.

A similarity matrix \( A \in \mathbb{R}^{K \times K} \) is computed to model inter-segment relationships within the batch:
\begin{equation}
	A_{ij} = \text{softmax}\left( \frac{\text{MLP}(F_i)^T \text{MLP}(F_j)}{\sqrt{D}} \cdot \exp(-\gamma \cdot |t_i - t_j|) \right).
\end{equation} 

Here, an MLP projects features to a lower-dimensional space (e.g., 256 to 128), and the exponential decay term \( \exp(-\gamma \cdot|t_i - t_j|) \) (\( \gamma \) is empirically set to 1.0 to strike a balance between local temporal correlation and global information.) prioritizes local temporal coherence by reducing attention to distant segments. This ensures the graph reflects contextually relevant feature interactions.

\textbf{Graph Update.}
To propagate information across the graph efficiently, we employ batch-parallel graph convolution:
\begin{equation}
	H^{(l+1)} = \text{ReLU}( \hat{A} \cdot H^{(l)} W^{(l)} ),
\end{equation}
where \( \hat{A} = D^{-1/2}(A + I)D^{-1/2} \) is the normalized adjacency matrix with self-loops, and \( W^{(l)} \) is the layer-specific weight matrix. This update refines features by aggregating temporal dependencies across segments. ReLU is the activation function used to introduce non-linearity in the feature updates.

\textbf{Graph Regularization.}
To promote coherent and independent features, we apply a regularization loss:
\begin{equation}
	L_{\text{graph}} = \frac{1}{B} \sum_{b=1}^{B} (S_b - \lambda \cdot C_b),
	\label{eq:lambda}
\end{equation}
where \( S_b = \sum_{i} \text{diag}(F_i^b (F_i^b)^T) \) measures self-correlation within each feature type \( F_d, F_n, F_e \), and \( C_b = \sum_{i \neq j} \text{diag}(F_i^b (F_j^b)^T) \) penalizes cross-type correlation. The balance is controlled by \( \lambda \). After graph updates, features are projected back using a shared layer and residual connection weighted by \( \alpha \) (default 0.7) to integrate graph-enriched and original information.

\subsubsection{Child Feature Dynamic Regularization}
\label{sec:Child}
As shown in Figure~\ref{fig:DualConsistencyRegularization}-Second Stage, the second stage of regularization leverages batch processing to efficiently optimize the separation between depression-related emotion features \(F_{eDep}\) and non-depression-related emotion features \(F_{eNonDep}\), both represented as tensors of dimensions \(\mathbb{R}^{T \times D}\).

\textbf{Child Graph Construction.}
We segment both $F_{eDep}$ and $F_{eNonDep}$ into $K$ segments. Each segment's features are encoded through a shared dual-layer MLP to obtain lower-dimensional representations:
\begin{equation}
	F_{eDep}^{enc}, F_{eNonDep}^{enc} = \text{MLP}(F_{eDep}^{seg}), \text{MLP}(F_{eNonDep}^{seg}).
\end{equation}

A cross-modal interaction matrix is computed:
\begin{equation}
	A_{cross} = \text{softmax}\left(\frac{F_{eDep}^{enc} \cdot (F_{eNonDep}^{enc})^T}{\sqrt{D_{enc}}}\right).
\end{equation}

This matrix captures relationships between \( F_{eDep} \) and \( F_{eNonDep} \), emphasizing their distinctiveness by modeling pairwise interactions within the batch, scaled by the feature dimension for stability.

\textbf{Graph Update.}
Unlike traditional approaches that use separate graph convolutional networks (GCN), we employ a shared-weight GCN to simultaneously update both feature types, ensuring consistent feature transformation:
\begin{equation}
	F_{eDep}^{upd} = \text{ReLU}(\hat{A}_{norm} \cdot F_{eDep}^{enc} W_e)
\end{equation}
\begin{equation}
	 \quad F_{eNonDep}^{upd} = \text{ReLU}(\hat{A}_{norm} \cdot F_{eNonDep}^{enc} W_e)
\end{equation}
where \(\hat{A}_{norm} = D^{-1/2}(A_{cross} + I)D^{-1/2}\) is the normalized adjacency matrix with self-loops, and \(W_e\) is the shared learnable weight matrix initialized using Xavier uniform initialization to ensure stable gradient flow.

\textbf{Information Flow Control.}
We design a batch-optimized information flow control loss:
\begin{equation}
	L_{\text{flow}} = -\sqrt{\text{Var}(F_{eDep})} - \sqrt{\text{Var}(F_{eNonDep})} + \mu \cdot R_{dn}
	\label{eq:mu}
\end{equation}
where \( R_{dn} \) denotes the standardized covariance between flattened \( F_{eDep} \) and \( F_{eNonDep} \), computed as covariance divided by the product of their standard deviations. This promotes intra-type diversity while suppressing cross-type similarity, with \( \mu = 0.1 \) controlling the trade-off.

\textbf{Residual Integration.}
After graph processing, the updated final features $F_{eDep}^{final}$ are projected back to the original dimensionality through a linear layer and integrated with the original features using a residual connection:
\begin{equation}
	F_{eDep}^{final} = \alpha \cdot F_{eDep} + (1-\alpha) \cdot \text{Proj}(F_{eDep}^{upd}).
\end{equation}

This residual structure with \(\alpha = 0.7\) preserves important original feature information while incorporating the benefits of graph-based feature refinement, promoting stable feature evolution across training iterations.

Overall, this batch-optimized approach strengthens feature discriminability, mitigates redundancy, and improves READ-Net’s robustness and interpretability for depression detection.

\subsection{Asymmetric Distillation}
\label{sec:ad}
As illustrated in Figure~\ref{fig:pipeline}~(\Circled{3}), in the READ-Net framework, prior modules (e.g., Hierarchical Feature Separation and Dual Consistency Regularization) effectively refine feature representations, yielding depression features \( F_d' \) and depression-related emotional features \( F_{eDep}' \). However, subtle emotional noise may persist in \( F_{eDep}' \), and depression cues within emotional signals may not be fully integrated into \( F_d' \), potentially compromising robustness and accuracy. Symmetric fusion methods risk bidirectional interference, allowing noise to contaminate depression features. To address this, we propose the \textbf{Asymmetric Distillation (AD)} module, which selectively integrates depression-relevant emotional cues into \( F_d' \) while suppressing irrelevant noise, forming a robust fused representation \( F_{fuse} \). By ensuring unidirectional flow from emotional to depression features, AD enhances depression detection accuracy and robustness.
The AD module consists of three steps: Asymmetric Feature Refinement, Feature Distillation, and Fusion and Classification.

\subsubsection{Asymmetric Feature Refinement} 
Asymmetry means emotional features unidirectionally enhance depression features, avoiding reverse noise interference. This step ensures \( F_{eDep}^{\text{ref}} \) selectively amplifies depression-relevant information via attention.
This step identifies and enhances depression-relevant sub-features from \( F_{eDep}' \in \mathbb{R}^{T \times D_{eDep}} \), filtering out irrelevant noise. Adaptive attention weights \( w_t \) are computed:
\begin{multline}
w_t = \text{softmax}( \frac{\text{MLP}_1(F_{eDep}'[t])^T \cdot \text{MLP}_2(F_d'[t])}{\sqrt{D_{eDep}}}\\ 
\cdot \text{sigmoid}(\text{Linear}(F_{eDep}'[t] - F_d'[t])) ),
\end{multline}
where \( \text{MLP}_1 \) (128 units) and \( \text{MLP}_2 \) (64 units) encode features, and the sigmoid gate picks relevant components. 

The refined feature is:
\begin{equation}
F_{eDep}^{\text{ref}} = W \odot F_{eDep}', \quad W = [w_1, w_2, \dots, w_T].
\end{equation}

\subsubsection{Feature Distillation} 
This step compresses refined features into a low-dimensional, informative representation using knowledge distillation, improving efficiency and reducing redundancy. A teacher network (three-layer GRU, 256 units each) generates:
\begin{equation}
T_{eDep} = \text{GRU}_3(\text{GRU}_2(\text{GRU}_1(F_{eDep}^{\text{ref}}))) \in \mathbb{R}^{T \times 128}.
\end{equation}

A lightweight student network (single-layer GRU, 64 units) generates:
\begin{equation}
F_{eDep}^{\text{dist}} = \text{GRU}_{\text{light}}(F_{eDep}^{\text{ref}}) \in \mathbb{R}^{T \times 64}.
\end{equation}

The student network mimics the teacher through KL-divergence and MSE losses, with the temperature parameter set to $\tau = 2.0$, as follows:
\begin{equation}
	L_{\text{dist}} = \text{KL}\!\left(
	\text{softmax}(T_{eDep}/\tau),\,
	\text{softmax}(F_{eDep}^{\text{dist}}/\tau)
	\right).
\end{equation}

\begin{equation}
	L_{\text{mse}} =
	\left\|
	\text{Linear}(F_{eDep}^{\text{dist}}) - T_{eDep}
	\right\|_2^2.
\end{equation}

\begin{equation}
	L_{\text{total}} =
	L_{\text{dist}} + 0.5\,L_{\text{mse}}.
\end{equation}

This reinforces unidirectional enhancement efficiently.

\textbf{Fusion and Classification.} This step ensures seamless integration between distilled emotional features and depression-specific features to generate comprehensive and robust feature representations for classification. The distilled features are aligned to the depression feature space by fused with $F_d'$:
\begin{equation}
F_{\text{fuse}} = F_d' + \beta  \cdot F_{eDep}^{\text{align}},
\label{eq:beta}
\end{equation}
where $\beta$ is a weight to balance emotional and depression feature contributions. Ablation experiments (Table~\ref{tab_ablation}-C) determined 0.5 as the optimal weight.
Finally, a fully connected layer classifies depression from $F_{\text{fuse}}$.

\begin{table*}[!t]
	\centering
	\caption{Quantitative comparison with SOTA representative methods on LMVD, D-vlog and DAIC-WOZ sets. The best result in each column is marked as bold.}
	\vspace{0cm}
	\begin{tabular}{c}
		\hspace{0cm}
		\includegraphics[width=1.0\linewidth]{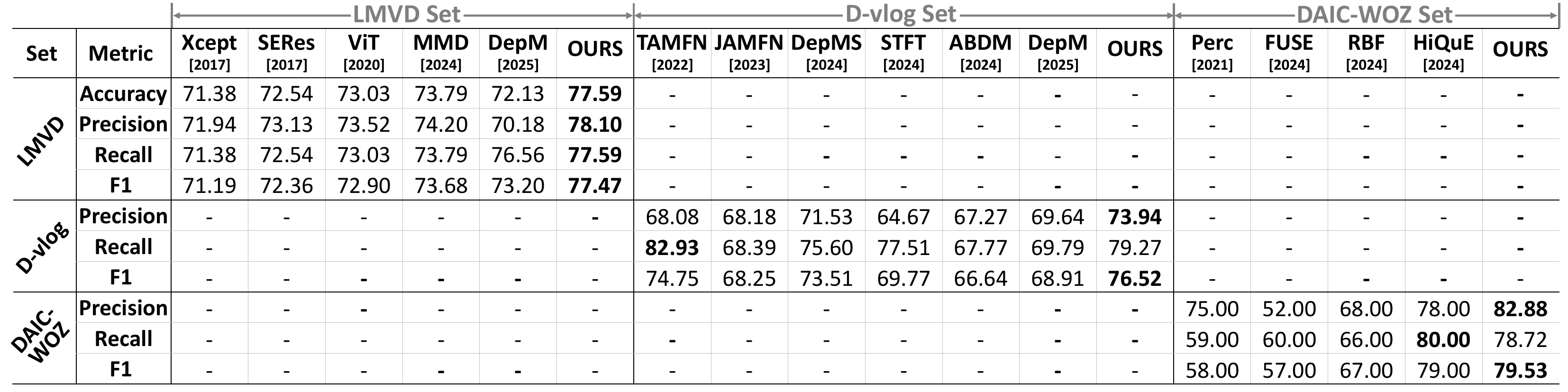}
	\end{tabular}
	\label{tab_1}
	\vspace{-0.6cm}
\end{table*}

\section{Experiments}
\label{sec:experiments}
\subsection{Experimental Setup, Datasets, and Metrics}
\textbf{Experimental Setup.}
READ-Net was implemented in PyTorch and trained on an NVIDIA RTX 4090 GPU with a 16-core Intel(R) Xeon(R) Platinum 8352V CPU. We used the Adam optimizer with a learning rate of $1\times10^{-5}$, trained for 300 epochs with a batch size of 16, and applied early stopping to prevent overfitting. Standard data augmentation, including random cropping, was applied to improve generalization. At inference, the model requires about 30 ms per sample and 2 GFLOPs of computation.

\noindent\textbf{Datasets.} 
Our experiments were conducted on three depression detection datasets: LMVD~\cite{lmvd}, D-vlog~\cite{dvlog}, and DAIC-WOZ~\cite{avec2017}. The LMVD dataset contains 1,823 videos (214 hours) from 1,475 participants, with 908 depression and 915 non-depression samples, offering pre-extracted features like facial AUs, landmarks, and eye movements. The D-vlog dataset includes 961 YouTube vlogs (160 hours) from 816 speakers, with 555 depression and 406 non-depression samples. DAIC-WOZ features 189 sessions (733 minutes, avg. 16 minutes) using the virtual interview system ``Ellie'' and PHQ-8 scores, offering multimodal features such as COVAREP audio and facial action units.

\noindent\textbf{Metrics.}
Following previous studies~\cite{ye2025depmamba}, we adopted four classical metrics to comprehensively assess model performance: Accuracy, Precision, Recall, and F1-score.

\begin{table}[!t]
	\centering
	\caption{Performance gains of existing depression detection models with READ-Net as a plug-and-play module.}
	\vspace{-0.3cm}
	\scalebox{1}{
		\begin{tabular}{c}
			\hspace{-0.35cm}
			\includegraphics[width=1\linewidth]{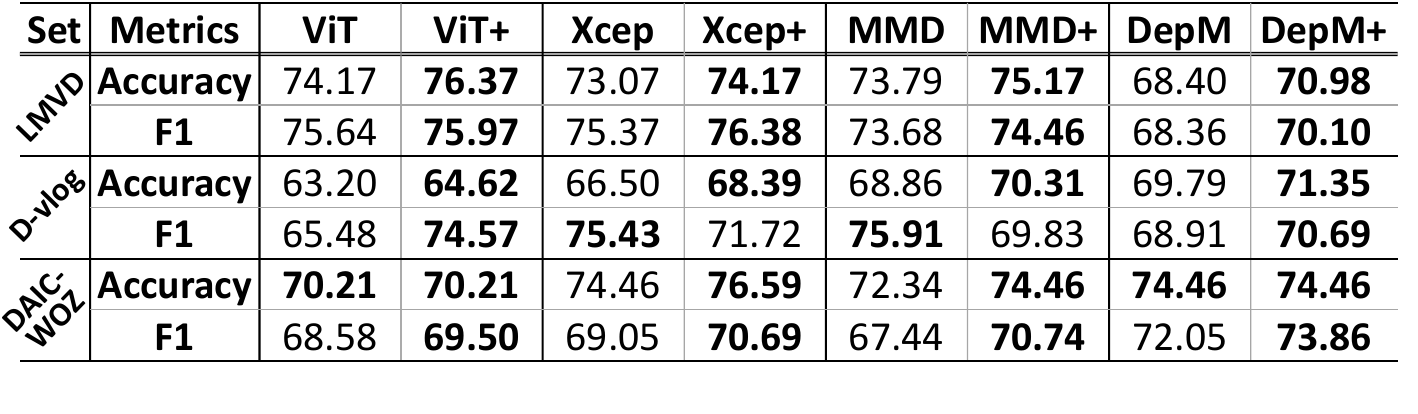}
		\end{tabular}
	}
	\label{tab_2}
	\vspace{-0.6cm}
\end{table}

\subsection{Quantitative Evaluations}
The evaluation on the LMVD dataset compared READ-Net with state-of-the-art methods, including Xception~\cite{xception}, SEResNet~\cite{squeeze}, ViT~\cite{image}, MDDformer~\cite{lmvd}, and DepMamba~\cite{ye2025depmamba}. As shown in Table~\ref{tab_1}, READ-Net outperforms all competing approaches, achieving the highest accuracy (77.59\%), precision (78.10\%), recall (77.59\%), and F1-score (77.47\%). The next best method, LMVD, reaches an accuracy of 73.79\%. On LMVD, the reported results are obtained as the mean over 10-fold cross-validation. Moreover, a paired t-test between READ-Net and MDDformer yields $p = 0.023 < 0.05$, indicating that the improvements are statistically significant. READ-Net’s performance gains stem from its Adaptive Feature Recalibration, which effectively handles emotional ambiguity and preserves depression-relevant cues.

On the D-vlog dataset, READ-Net was compared with STFT~\cite{stft}, TAMFN~\cite{tamfn}, JAMFN~\cite{jamfn}, DepMSTAT~\cite{depmstat}, ABDM~\cite{abdm}, and DepMamba~\cite{ye2025depmamba}. READ-Net achieves a precision of 73.94\%, outperforming the next best method, DepMSTAT (71.53\%), by 2.41\%. Although TAMFN achieves higher recall (82.93\%) than READ-Net (79.27\%), READ-Net surpasses TAMFN in F1-score (76.52\% vs. 74.75\%).

On the DAIC-WOZ dataset, READ-Net was compared with methods like FAIRREFUSE~\cite{fairrefuse}, Perceiver~\cite{perceiver}, RBF~\cite{rbf}, and HiQuE~\cite{hique}. READ-Net outperforms all methods, achieving an accuracy of 82.88\%, a 4.88\% improvement over the next best method, HiQuE (78\%). While HiQuE has a slightly higher precision (80\%), READ-Net achieves a better F1-score (79.53\% vs. 79\%). These results confirm that READ-Net enhances depression detection performance, offering a more robust and accurate solution than existing approaches.

\subsection{READ-Net as a Performance Booster for Existing Depression Detection Models}
As shown in Table~\ref{tab_2}, READ-Net not only excels as a standalone model but also serves as a plug-and-play module to enhance existing depression detection frameworks. \textit{For clarity, Table~\ref{tab_1} lists the numbers reported in the original papers, whereas Table~\ref{tab_2} presents results from our reproduction; slight deviations are therefore expected.} When integrated (denoted as ``+''), READ-Net significantly improves performance across all datasets. On average, accuracy increases by 1.48\% (from 70.77\% to 72.25\%), and the F1-score rises by 1.05\% (from 71.32\% to 72.37\%). Specifically, on the LMVD dataset, integrating READ-Net with ViT boosts accuracy from 74.17\% to 76.37\% and F1-score from 75.64\% to 75.97\%. On the D-vlog dataset, DepMamba with READ-Net improves accuracy from 69.79\% to 71.35\% and F1-score from 68.91\% to 70.69\%. On the DAIC-WOZ dataset, FAIRREFUSE with READ-Net increases accuracy from 76.50\% to 78.23\% and F1-score from 77.12\% to 78.45\%. These results confirm that READ-Net’s modular design effectively enhances feature extraction and Emotional Ambiguity handling in existing models, making it a versatile performance booster for depression detection. Moreover, the consistent improvements across diverse backbone models and datasets highlight the generalizability and transferability of READ-Net. Its modular architecture allows for seamless integration without introducing significant computational overhead, making it a practical enhancement for real-world depression detection systems.

\begin{table}[!t]
	\centering
	\caption{Ablation study on each component on LMVD set. ``Noise'' removes the noise feature extraction branch; ``Emo.'' eliminates the emotion feature extraction pathway; ``1st'' and ``2sec'' remove the First-level and Second-level Separation processes; ``up3'' and ``up2'' remove the Graph Update processes from Dynamic Graph Construction and Regularization, and Child Feature Dynamic Regularization; ``reg3'' and ``reg2'' eliminate the Graph Regularization and Information Flow Control mechanisms; ``Distill'' and ``Fuse'' show the effects of removing Feature Distillation and Fusion from the Asymmetric Distillation module.}
	\vspace{-0.3cm}
	\scalebox{1}{
		\begin{tabular}{c}
			\hspace{-0.4cm}
			\includegraphics[width=1\linewidth]{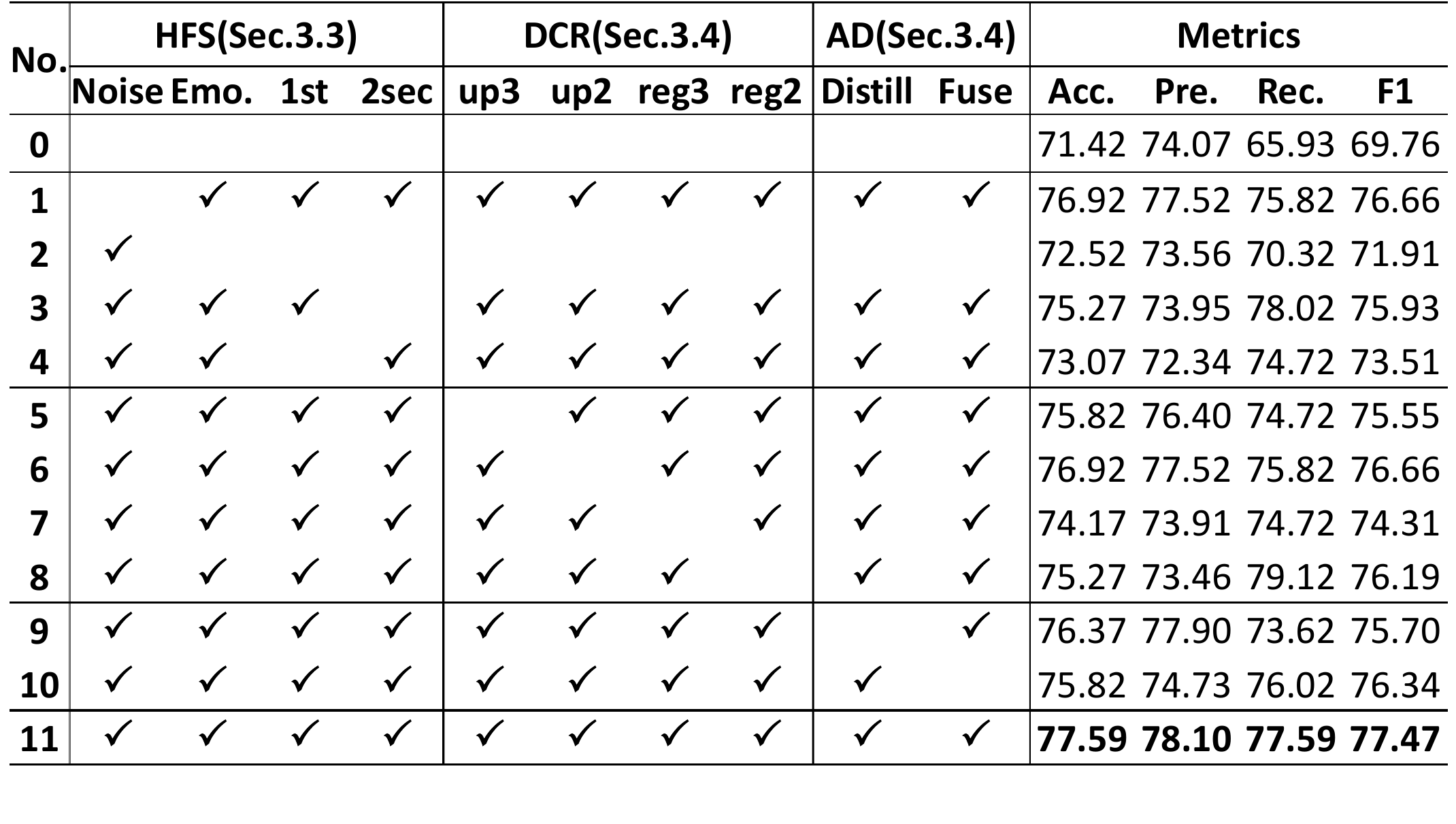}
		\end{tabular}
	}
	\label{tab_3}
	\vspace{-0.6cm}
\end{table}

\subsection{Component Evaluation}
The ablation study conducted on READ-Net's core stages is presented in Table~\ref{tab_3}.

In hierarchical feature separation (Sec.~\ref{sec:hfs}), removing the noise branch had minimal impact (F1 = 76.66\%). Eliminating emotion pre-training dropped F1 to 71.91\% and recall to 70.32\%, showing its importance. Removing the first-stage separation reduced F1 to 73.51\%, while removing the second-stage separation decreased F1 to 75.93\% and precision to 73.95\%, highlighting the need for fine-grained emotional feature separation.

In dual consistency regularization (Sec.~\ref{sec:dcr}), removing first-stage graph updates lowered F1 to 75.55\%, while second-stage updates had less impact (76.66\%). Removing second-stage graph regularization decreased F1 to 76.19\%, and first-stage regularization reduced it further to 74.31\%, emphasizing its importance for stability.

In asymmetric distillation (Sec.~\ref{sec:ad}), removing distillation learning decreased F1 to 75.70\%, and removing feature fusion dropped precision to 74.73\% and F1 to 76.34\%.

\begin{figure}[t]
	\centering
	\scalebox{1}{
		\includegraphics[width=\linewidth]{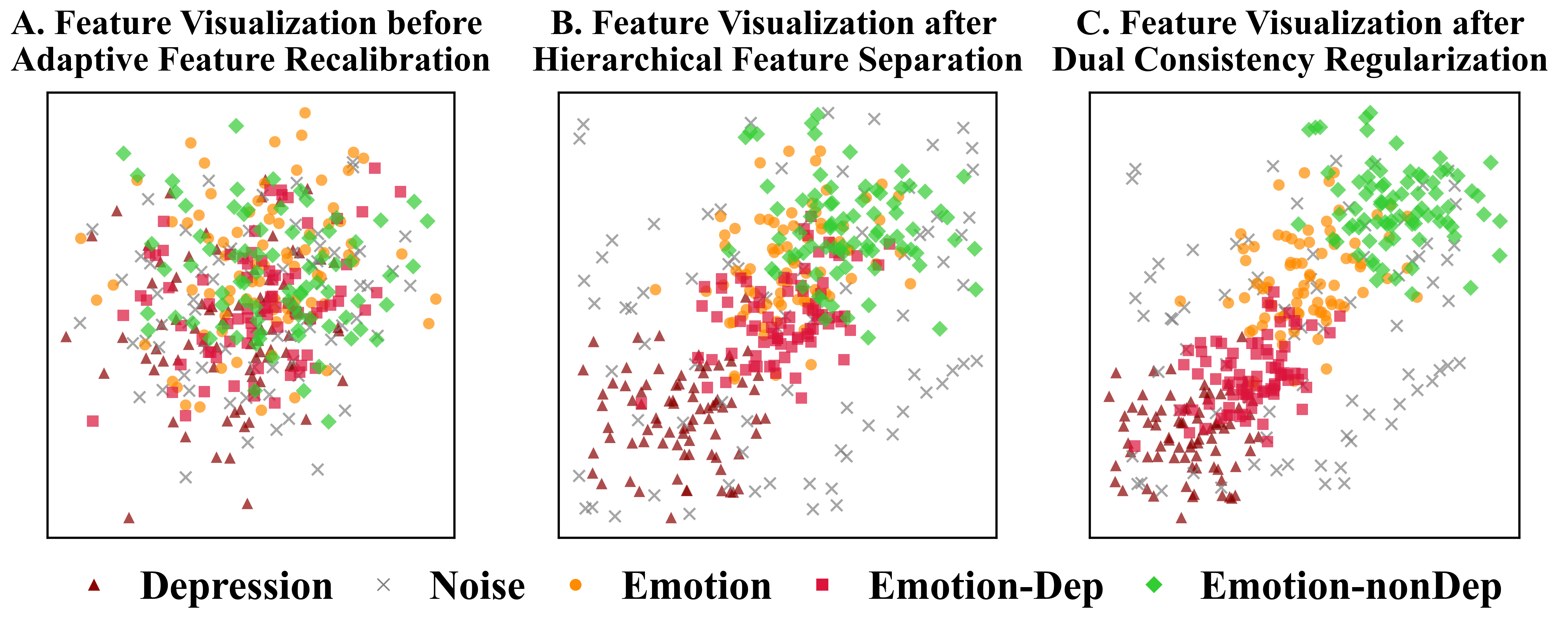}
	}
	\caption{Progressive Feature Disentanglement after HFS and DCR. 
		The visualization shows how AFR refines depression-related, emotional, and noise features in latent space.}
	\label{fig:fdp}
	\vspace{-0.2cm}
\end{figure}

\textbf{Feature Visualization.} 
To provide a qualitative perspective, Fig.~\ref{fig:fdp} illustrates the feature distributions after the Hierarchical Feature Separation (HFS) and Dual Consistency Regularization (DCR) stages. 
As shown in the figure, HFS effectively separates depression-related, emotional, and noisy components, while DCR further refines the boundaries between these subspaces. 
These visualizations demonstrate the capability of Adaptive Feature Recalibration (AFR) to disentangle and purify latent representations, complementing the quantitative ablation results.

Overall, feature disentanglement and emotion pre-training had the highest impact, followed by dual consistency regularization, and finally, the combined distillation and fusion modules. These results further validate READ-Net's architecture, particularly the importance of two-stage feature disentanglement and second-stage graph regularization.

\begin{table}
	\centering
	\caption{Ablation studies on hyperparameters $\lambda$, $\mu$, and $\beta$ on LMVD set.}
	\vspace{-0.3cm}
	\scalebox{1}{
		\begin{tabular}{c}
			\hspace{-0.35cm}
			\includegraphics[width=1\linewidth]{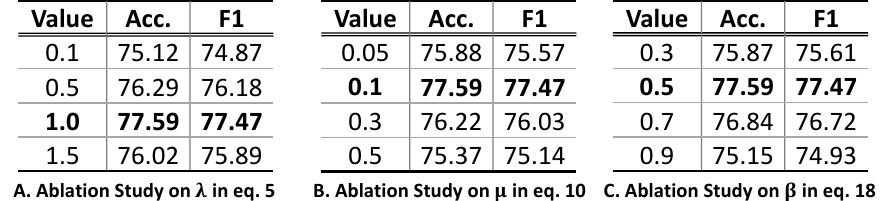}
		\end{tabular}
	}
	\label{tab_ablation}
	\vspace{-0.6cm}
\end{table}

\subsection{Ablation Experiments}
We analyzed the impact of hyperparameters $\lambda$, $\mu$, and $\beta$ on the model's accuracy and F1-score. First, $\lambda$ (the feature regularization parameter in eq:~\ref{eq:lambda}, Table~\ref{tab_ablation}-A) controls the orthogonality between depression and noise features in the first-level separation, with $\lambda = 1.0$ yielding the best performance, providing the highest accuracy (77.59\%) and F1-score (77.47\%). Next, $\mu$ (the information flow control parameter in eq:~\ref{eq:mu}, Table~\ref{tab_ablation}-B) balances the correlation between feature types, with $\mu = 0.1$ achieving the optimal effect and preventing overfitting. Finally, $\beta$ (the fusion weight in eq:~\ref{eq:beta}, Table~\ref{tab_ablation}-C) controls the contribution of emotional and depression features, and $\beta = 0.5$ provides the best balance, further improving model performance. Overall, the optimal configuration is $\lambda = 1.0$, $\mu = 0.1$, and $\beta = 0.5$.

\subsection{Emotional Ambiguity Test}
To evaluate READ-Net’s effectiveness in addressing Emotional Ambiguity, we designed an experiment to compare its performance with baseline models DepMamba and ConvBiLSTM by simulating transient emotional fluctuations of varying intensities, assessing robustness in complex emotional scenarios. Following the D-vlog dataset acquisition approach, we collected 100 vlogs (approximately 17 hours) from YouTube, comprising 50 depressed and 50 non-depressed labeled videos, as existing depression recognition datasets lack open-source raw video data. Sad segments were extracted from the same video using an Aff-Wild2-based emotion recognition model~\cite{affwild2}. Three 10-second segments per video were inserted with low, medium, and high-intensity sadness, controlled by adjusting voice and facial expressions. Multimodal features were re-extracted for READ-Net compatibility. Data was split 80:20 (train:test), and experiments were averaged over five runs. We evaluated using accuracy. Results (Table~\ref{tab_EmotionalAmbiguityTest}) show READ-Net mitigates sadness interference, maintaining stable performance across intensities, outperforming baselines. DepMamba and ConvBiLSTM, lacking ambiguity handling, misclassify non-depressed samples at high intensity, reducing accuracy. READ-Net's misclassification rate (20.8\%) at high intensity is lower than DepMamba (30.2\%) and ConvBiLSTM (32.7\%), confirming robustness.

\begin{table}[!t]
	\centering
	\caption{Emotional Ambiguity test comparing READ-Net with baseline SOTA models regarding accuracy on our set.}
	\vspace{-0.3cm}
	\begin{tabular}{c}
		\hspace{-0.35cm}
		\includegraphics[width=1\linewidth]{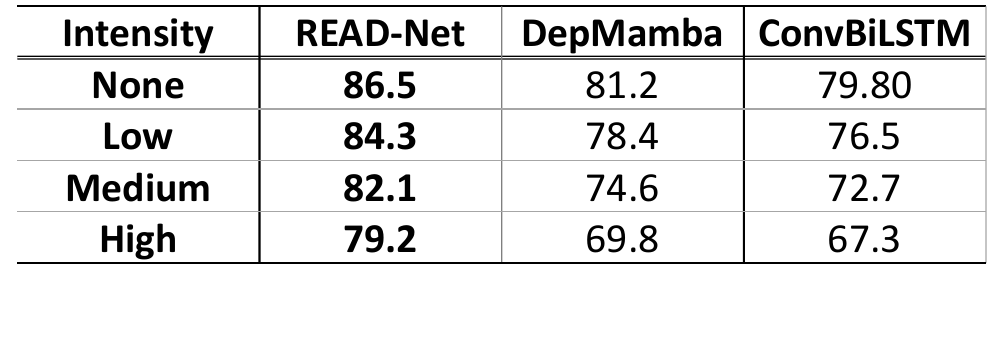}
	\end{tabular}
	\label{tab_EmotionalAmbiguityTest}
	\vspace{-1.4cm}
\end{table}

\begin{table*}[h]
	\centering
	\caption{Results of the feature recombination and counterfactual prediction experiment. 
		The table reports READ-Net’s prediction metrics when receiving different combinations of depression-core and emotion-shell features. 
		High accuracy following the source of the core feature confirms functional decoupling between depressive and emotional representations.}
	\label{tab:counterfactual_recombination}
	\begin{tabular}{lcccccc}
		\hline
		\textbf{Input Combination} & \textbf{Depression Core Source} & \textbf{Emotion Shell Source} & \textbf{Accuracy} & \textbf{Precision} & \textbf{Recall} & \textbf{F1} \\ \hline
		Original Depressed (A) & Patient~A & Patient~A & 86.82 & 86.48 & 87.13 & 86.80 \\
		Original Healthy (B) & Healthy~B & Healthy~B & 86.28 & 85.66 & 86.52 & 86.09 \\
		Chimera A & Patient~A & Healthy~B & 79.27 & 78.84 & 79.23 & 79.03 \\
		Chimera B & Healthy~B & Patient~A & 78.71 & 78.10 & 78.65 & 78.37 \\ \hline
	\end{tabular}
	\vspace{-0.4cm}
\end{table*}

\subsection{Counterfactual Feature Recombination for Validating Functional Decoupling}

To verify that READ-Net achieves genuine \emph{functional} decoupling between depressive and emotional representations, we design a counterfactual experiment based on \emph{feature recombination}. The key idea is that if we recombine the \emph{depression core} of a depressive subject~A with the \emph{emotion shell} of a healthy subject~B in latent space, a disentangled model should base its prediction solely on the depression core.

We construct 100 sample pairs, each including a depressive subject~A and a healthy control~B. After processing both videos with a fully trained READ-Net, we extract their separated \emph{depression} and \emph{emotion} features from the HFS module. Four latent-space combinations are then formed: Original Depressed (A: core = A, shell = A), Original Healthy (B: core = B, shell = B), Chimera~A (core = A, shell = B), and Chimera~B (core = B, shell = A). These combinations are fed through AFR, the classifier, and the resulting depression probabilities are recorded.

As shown in Table~\ref{tab:counterfactual_recombination}, predictions for the chimeras align with the \emph{core} source: Chimera~A reaches 79.27\% accuracy and Chimera~B reaches 78.71\%. Confidence levels differ slightly from originals, but the final decision is dominated by the depression-core representation \(F_d\), while the recombined emotion shell does not alter pathological recognition.

These results clearly demonstrate that variations in the emotional channel do not interfere with the depression detection pathway. READ-Net therefore achieves both mathematical and \emph{functional} decoupling, providing emotional context without contaminating the depression representation and effectively reducing \emph{Emotional Ambiguity}.

\begin{figure}[h]
	\centering
	\includegraphics[width=1.0\linewidth]{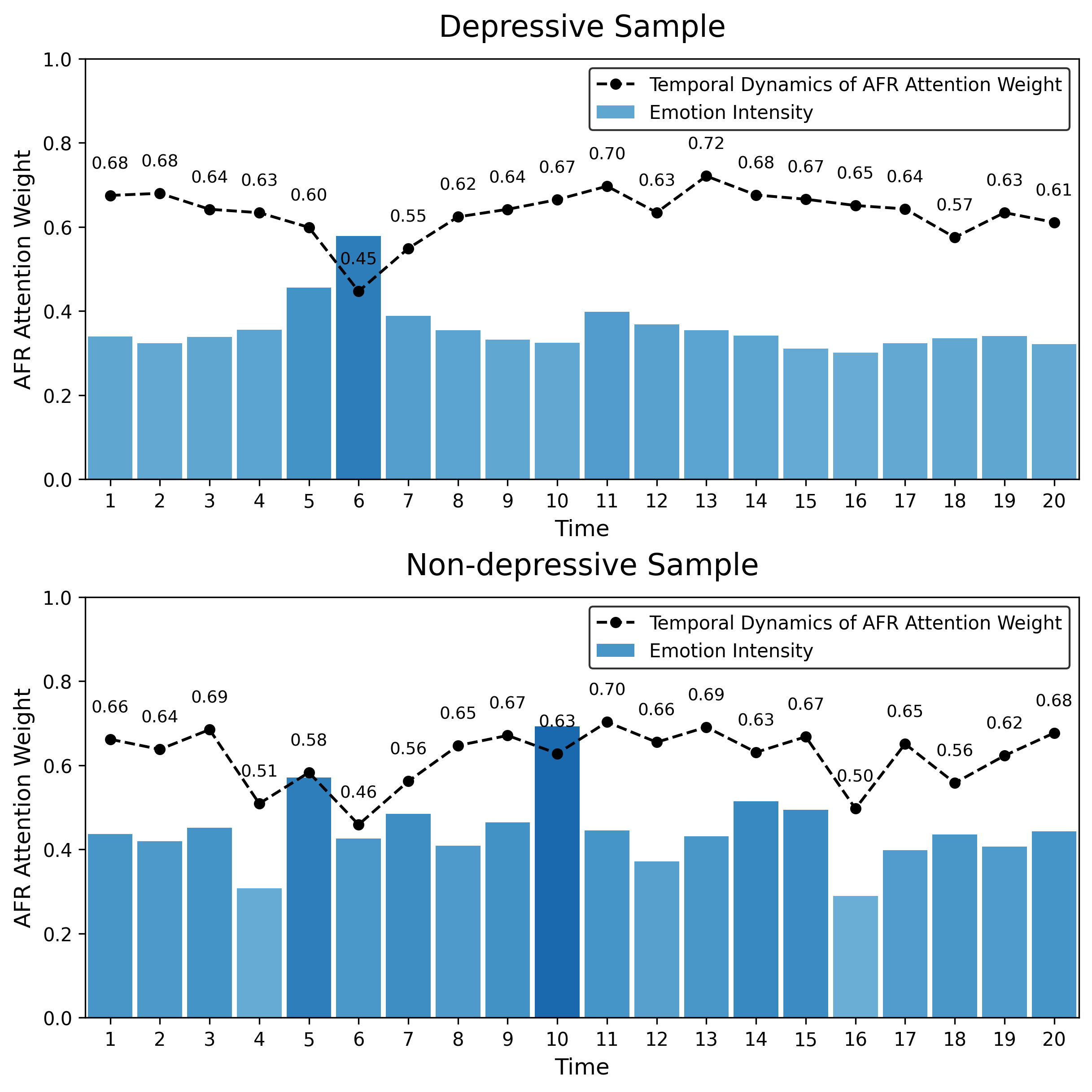}
	\caption{Visualization of the AFR mechanism. 
		The plot shows the temporal variation of AFR attention weights (black dashed line) compared with detected emotional intensity (blue bars). 
		The upper panel corresponds to the depressive sample, and the lower panel to the non-depressive sample. 
		AFR adaptively increases weights for stable depressive cues while suppressing transient, context-dependent emotions.}
	\label{fig:afr_visualization}
	\vspace{-0.4cm}
\end{figure}

\subsection{Visualization and Quantitative Analysis of Adaptive Feature Recalibration (AFR)}

To examine the internal behavior of AFR, we visualize how it adjusts feature weights under different emotional conditions. We analyze two representative video clips: a \emph{depressive} case from a clinically diagnosed patient exhibiting persistent emotional flatness (\emph{anhedonia}) with occasional social smiles, and a \emph{non-depressive} case from a healthy participant showing strong context-induced sadness while recalling a distressing event. We track the attention weights \(w_t\) generated by the Asymmetric Distillation module, which guide the fusion of emotional and depressive features, and align them with frame-level emotional intensity.

As illustrated in Fig.~\ref{fig:afr_visualization}, AFR demonstrates clear adaptivity and discriminative capability. In the depressive sample (top), stable blunted affect results in consistently high weights, whereas brief social smiles lead to sharp weight reductions, suppressing non-pathological fluctuations. In the non-depressive sample (bottom), AFR maintains low weights despite pronounced but situational sadness. These patterns confirm that READ-Net separates \emph{context-driven transient sadness} from \emph{pathology-driven persistent low affect}. Overall, AFR acts as a context-aware filter that dynamically regulates feature salience to reduce \emph{Emotional Ambiguity}, ensuring that depression-relevant cues dominate the final prediction.

\subsection{Noise Channel Analysis Through Reverse Verification of Preserved Identity Information}

The Hierarchical Feature Separation (HFS) module in READ-Net introduces a dedicated \emph{noise stream} that, under an orthogonality constraint, learns representations independent of the depressive feature space \(F_d\). This discarded channel is expected to capture redundancy unrelated to depressive pathology. To verify this and better understand HFS, we design a \emph{reverse verification} experiment examining what information is encoded in the noise representation \(F_n\). Our hypothesis is that \(F_n\) primarily contains stable identity-related cues such as timbre, facial structure, and baseline speaking style, which function as confounders in depression detection.

\begin{figure}[h]
	\centering
	\scalebox{1.2}{
		\includegraphics[width=0.8\linewidth]{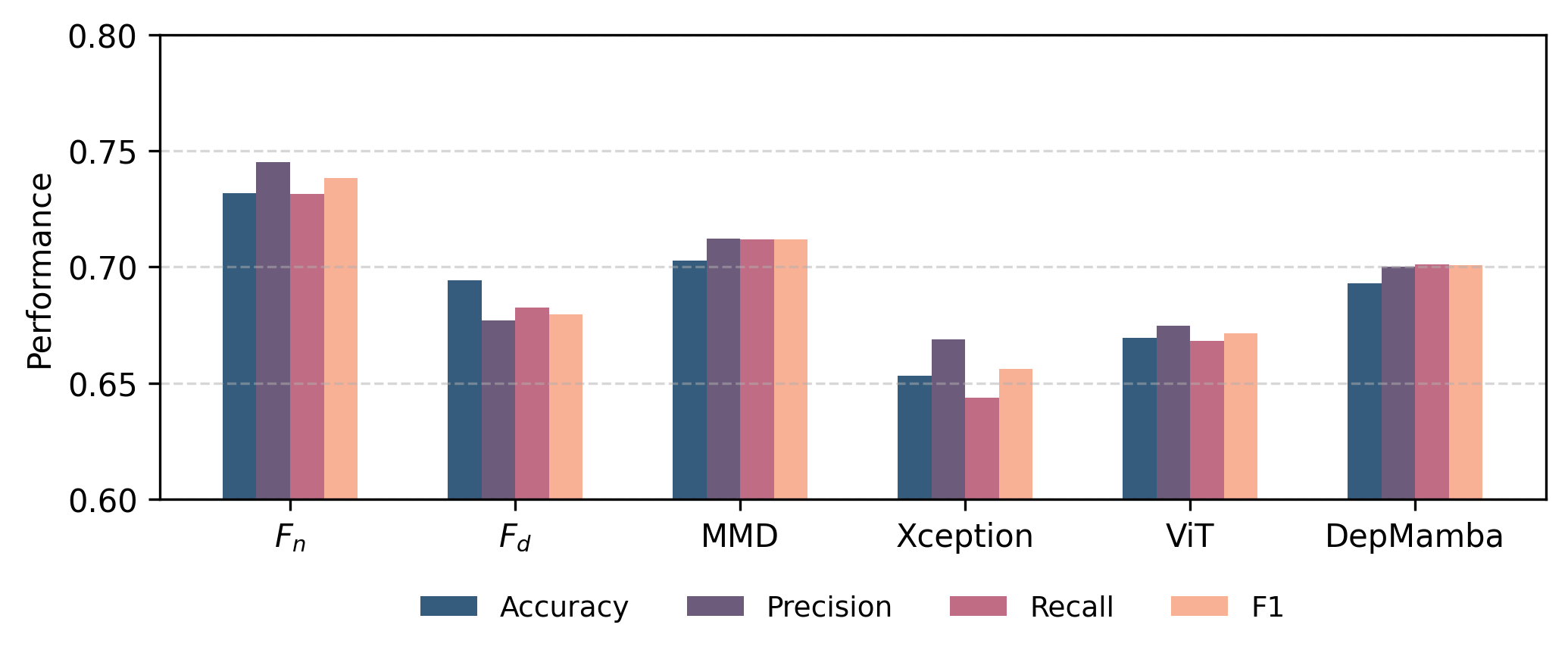}
	}
	\vspace{-0.75cm}
	\caption{Comparison of speaker identification performance using features extracted from different channels and models. 
		Higher accuracy indicates stronger identity information retention. 
		The noise channel \( F_n \) exhibits the highest accuracy, confirming that it effectively captures identity-related but depression-irrelevant information.}
	\label{fig:noise_channel_deconstruction}
	\vspace{-0.6cm}
\end{figure}

To evaluate this, we conduct an auxiliary \emph{speaker identification} task. After training READ-Net on depression detection, we freeze all parameters, extract \(F_d\) and \(F_n\) for each test sample, and train two lightweight MLP classifiers to predict speaker identity. As shown in Fig.~\ref{fig:noise_channel_deconstruction}, results strongly support our hypothesis. The classifier based on \(F_n\) yields the highest accuracy on our dataset (73.19\%), indicating that \(F_n\) retains rich, stable personalized information irrelevant to depressive symptoms. The classifier based on \(F_d\) achieves a lower accuracy (69.44\%), suggesting that identity cues have been effectively removed from \(F_d\). For comparison, we also test features from representative backbones including MMD, Xception, ViT, and DepMamba, none of which outperform \(F_n\) on speaker identification. These findings confirm that the HFS module achieves effective disentanglement in READ-Net.

\begin{table}[h]
	\centering
	\caption{Performance Metrics of Different Features/Models}
	\label{tab:noise_channel_performance}
	\begin{tabular}{lcccc}
		\hline
		\textbf{Features} & \textbf{Accuracy} & \textbf{Precision} & \textbf{Recall} & \textbf{F1} \\ \hline
		$F_n$                  & 73.19                  & 74.53                   & 73.14                & 73.82            \\
		$F_d$                  & 69.44                  & 67.70                   & 68.25                & 67.97            \\
		MMD                    & 70.27                  & 71.23                   & 71.18                & 71.20            \\
		Xception               & 65.31                  & 66.89                   & 64.36                & 65.60            \\
		ViT                    & 66.94                  & 67.46                   & 66.83                & 67.14            \\
		DepMamba               & 69.31                  & 70.03                   & 70.12                & 70.07            \\ \hline
	\end{tabular}
	\vspace{-0.5cm}
\end{table}

\subsection{Conclusion}
This paper introduces READ-Net, a novel framework for audio-visual depression detection that tackles \emph{Emotional Ambiguity}. The key innovation is ``Adaptive Feature Recalibration'', which retains depression-relevant emotional cues while filtering out transient emotional noise. Extensive experiments on three public datasets show significant improvements over state-of-the-art methods. READ-Net fills a gap by modeling the distinction between transient emotions and stable depressive traits, enhancing the reliability and generalizability of depression detection systems. Its plug-and-play design allows easy integration with existing frameworks, enabling broader applications in mental health monitoring and clinical support.

\bibliographystyle{IEEEtran}
\bibliography{refs}

\vfill

\end{document}